# TI-Capsule: Capsule Network for Stock Exchange Prediction

Ramin Mousa
Department of Computer Science and
Technology, Zanjan University, Iran
Raminmousa@znu.ac.ir

Sara Nazari
Department of Computer Engineering
Arak branch, Islamic Azad
Markazi, Iran
S-nazari@iau-arak.ac.ir

Ali Karhe Abadi
Department of Computer Engineering
University of Tehran
Tehran, Iran
a.karkehabadi@ut.ac.ir

Reza Shoukhcheshm
Department of Computer Science
University of Amirkabir, Tehran, Iran
Tehran, Iran
Reza.shookhcheshm.71@gmail.com

Mohammad Niknam Pirzadeh
Department of Computer Engineering
Arak branch, Islamic Azad
Markazi, Iran
nikimail2004@gmail.com

Leila Safari
Department of Computer Science
University of Zanjan
Zanjan, Iran
lsafari@znu.ac.ir

A B S T R A C

Today, the use of social networking data has attracted a lot of academic and commercial attention in predicting the stock market. In most studies in this area, the sentiment analysis of the content of user posts on social networks is used to predict market fluctuations. Predicting stock marketing is challenging because of the variables involved. In the short run, the market behaves like a voting machine, but in the long run, it acts like a weighing machine. The purpose of this study is to predict EUR/USD stock behavior using Capsule Network on finance texts and Candlestick images. One of the most important features of Capsule Network is the maintenance of features in a vector, which also takes into account the space between features. The proposed model, TI-Capsule (Text and Image information based Capsule Neural Network), is trained with both the text and image information simultaneously. Extensive experiments carried on the collected dataset have demonstrated the effectiveness of TI-Capsule in solving the stock exchange prediction problem with 91% accuracy.

**Introduction:**

Social networks such as Twitter has revolutionized the user's communication methods by expanding the use of virtual social networks and sharing a lot of information through users, the use of information on these social networks and their analysis were considered by researchers. Many researchers and analysts have used the Twitter social network as a rich source of information to predict the stock market (Bollen, Mao, and Zeng 2011; Zhang, Fuehres, and Gloor 2011; Li, Zhou, and Liu 2016; Hamed, Qiu, and Li 2015; Bartov, Faurel, and Mohanram 2018; Zhang 2013; Rao and Srivastava 2012). The question that many types of research have been seeking to answer is whether the analysis of data from these social networks can assess the state of the society or even predict the upcoming events?

Many people on the web are affected by the comments. This is especially true for product visits, which shows that people are influenced by shopping behavior. In addition, the information provided by people on the web is considered more reliable than the information provided by the seller. From the perspective of manufacturers, every person is a potential customer. Therefore, knowing what they like or dislike can help in creating new products (Van Kleef, Van Trijp, and Luning 2005) as well as managing and improving existing products (Pang and Lee 2008). In addition, understanding the relationship between the information available in customer reviews of products and the information provided by companies helps to benefit from and use these reviews to improve sales (Chen and Xie 2008). Sentiment analysis (SA) as an auxiliary factor has played an important role in stock forecasting and has been used in many studies (Zhao et al. 2016). SA of texts has been of interest to researchers in recent years (Ravi and Ravi 2015; Kharde and Sonawane 2016; Maerz and Puschmann 2020). SA means discovering and recognizing people's positive or negative feelings about a problem or product in the texts (Zucco et al. 2020).

The purpose of this study is to predict the increase/decrease of the EUR/USD stock index using the visual, numerical, and contextual information extracted from the stock and Twitter blogs. Many types of research in the literature have used the term stock sentiment(de O. Carosia, Coelho, and Silva 2019; Zadrozny 2019; Kim, Ryu, and Yang 2019) for this problem, so in the present study, the stock sentiment term is used instead of SA, which means rising prices indicate bullish market sentiment while falling prices indicate bearish market sentiment. Since this analysis deploys information such as metadata, contextual data, and visual data in relation to these stock market, the details of these data will be discussed in the present study.

As deep learning-based techniques have shown great success in many areas in recent years, there is a hope to resolve the limitations of existing techniques in stock forecasting using deep learning-based approaches, thereby improving the accuracy of forecasting future stock status. One of the most important constraints that deep learning can overcome is the possibility of incorporating diverse and predictive features, which leads to rich training and improved feature engineering in model learning. In contrast, using traditional machine learning techniques, the model accuracy decreases with increasing the number and variety of features, as they are only able to model simple and nonlinear relationships among data [9,10]. So, the benefits of deep learning can be exploited and integrated with the characteristics of the sentiment analysis of registered users' opinions and reviews (as input into the stock market forecasting model), which will hopefully improve forecasting accuracy in stock prediction.

Multi-modal SA is a Machine Learning (ML) problem that has attracted much attention in the last decade (Sun et al. 2019; Zhang et al. 2019; Shad Akhtar et al. 2019). The lack of natural structure in natural language has made the field of SA challenging. In the literature, the authors use the multi-modal state to compensate for incomplete information about learning models in SA. Combine textual, visual, and audio data to train learning models in Multi-modal SA is common(Kaur and Kautish 2019). In this study, the opinions that users post on the web (Tweeter), the finance news (Investing site), the finance features (forex factory), and the Candlestick charts (Windsor brokers) have been deployed for making decisions on stock sentiment prediction. In this research, we first look at the past efforts and then look for a way to classify stock sentiment based on the relationship between the user's tweets and stock price changes in different time intervals. Then, we propose a method for predicting the stock market sentiment in different length intervals based on text, features, and image information.

The paper is structured as follows. In Section 1, a survey of various approaches to stock market prediction is provided. Section 2 describes data collection and web mining process. Section 3 contains information on the materials used in the proposed approach, including the implementation tools, pre-trained word embeddings, and evaluation measures. Section 4 describes the model architecture, while the overall architecture of the TI-Capsule approach evaluated in Section 5. Finally, Section 6 concludes the article.

## 1. RELATED WORKS

The statistical, pattern recognition, machine learning, and hybrid approaches, together with sentiment analysis, have been used in many types of research to solve the problem of financial time series analysis. This section provides a brief review of some of the works in the literature that have used these techniques.

### 1.1 Statistical Approaches

Prior to the advent of machine learning techniques, statistical techniques were used to analyze time series. Numerous statistical approaches have been used in the literature to analyze the stock market and have obtained satisfactory results. The Exponential Smoothing Model (ESM) approach is one of the most important approaches of this type, which has been widely mentioned in the literature. This approach uses the exponential window function for smoothing time series data (Billah et al. 2006). Other approaches such as the Auto-Regressive Moving Average (ARMA) (Chen, Wang, and Huang 1995), the Auto-Regressive Integrated Moving Average (ARIMA) (Yu et al. 2014), the Generalized Autoregressive Conditional Heteroskedastic (GARCH)(Bollerslev 1986) volatility, and the Smooth Transition Autoregressive (STAR)(Teräsvirta, Van Dijk, and Medeiros 2005) model are also used for time series analysis. In some cases, these models have been able to compete with traditional machine learning approaches. For example,(De Faria et al. 2009) compared the Artificial Neural Network (ANN) and adaptive ESM models to predict stocks in Brazil. While the ANN model achieved a lower Root Means Square Error (RMSE), the ESM model was also acceptable and, in some cases, similar results, indicating the strength of these models.

### 1.2 Pattern Recognition

Pattern recognition techniques use previous (historical) patterns to predict future trends. They extract logical patterns for this purpose. The work of (Fu et al. 2005) was among the first to identify patterns of time that applied the concept of human visualization as well. The results of experiments [28] show that the approach does not only reduces the dimensions but also allows for early identification of patterns. (Parracho, Neves, and Horta 2010) proposed an approach by combining template matching with Genetic Algorithms (GA) to create an algorithmic trading system. Pattern matching has been used to identify ascending trends while GA has been deployed to optimize the parameters used in template matching. (Phetchanchai et al. 2010) proposed an innovative method for analyzing time series data, taking into account the zigzag motion in the data. The PIP technique has been chosen to identify zigzag motions, and the Zigzag based Mary tree (ZM-tree) was used to organize the points. The proposed method performs better than other methods, such as Specialized Binary Trees (SB-Tree). Examples of other methods that use Pattern Recognition can be found in the studies of Cervelló-Royo et al (Cervelló-Royo, Guijarro, and Michniuk 2015), Chen and Chen (Chen and Chen 2016), Arévalo et al. (Arévalo et al. 2017) and Kim et al. (Kim et al. 2018).

### 1.3 Machine Learning Approaches

Many machine learning approaches have been explored to predict stock prices. Artificial Neural Network (ANN) and Support Vector Regression (SVR) are the two most widely used approaches. These approaches are divided into supervised and unsupervised approaches:

- **Supervised approaches:** These approaches can provide meaningful analysis of stock prices based on historical prices. Bernal et al. (Bernal, Fok, and Pidaparthi 2012) implemented Echo State Networks (ESN), a subclass of Recurrent Neural Networks (RNN), to predict S&P 500 stock prices using the price, moving averages, and volume as features. This approach was tested and evaluated on 50 different stocks and achieved a test error of 0.0027. Various machine learning approaches have been used in (Vu et al. 2012), such as decision trees, and hybrid approaches to predict and classify Apple (AAPL), Google (GOOG), Microsoft (MSFT) and Amazon (AMZN) stocks. For this purpose, 5001460 tweets were collected and, according to the innovative features considered, achieved 82.93, 80.49, 75.61, and 75% accuracy for AAPL, GOOG, MSFT, and AMZN, respectively. Xu and Keelj (Xu and Keelj 2014) proposed a model for predicting stock changes for the next day, depending on the sentiment gained from March 13th, 2012-May 31st, 2012. Their proposed model consists of two basic phases. In the first phase, they attempted to classify the tweets as positive and negative using Natural Language processing (NLP) techniques. In the second phase, machine learning algorithms such as Decision Tree (DT), Support Vector Machine (SVM), and Naive Bayes (NB) were used to classify polar tweets. Balling et al (Ballings et al. 2015), tested ensemble approaches, including Random Forest,

AdaBoost, and Kernel Factory against single classifier models such as Neural Networks, Logistic Regression, Support Vector Machines, and K-Nearest Neighbor on data from 5767 European companies, which the Random Forest method gained a better result than other approaches. Other Supervised approaches such as Neural Network (NN) (Zadrozny 2019), Multiple Linear Regression (MLR) (Ariyo, Adewumi, and Ayo 2014), Supper Vector Regression (SVR) (Izzah et al. 2017), j48 algorithm (Ouahilal et al. 2016), LSTM and DNN (Kamble 2017), and LSTM (Shah, Campbell, and Zulkernine 2018) have also been used in the literature for this purpose.

- **Unsupervised approaches:** Unsupervised learning also applies when it is difficult to collect labeled data. This learning method basically can help to identify correlations in a non-correlated dataset such as stock market data. Dowell et al (Powell, Foo, and Weatherspoon 2008) compared the SVM and K-means on S and P 500 data using PCA (Principal Component Analysis) to reduce their dimensions and concluded that both methods have relatively similar performance. These algorithms achieved %89 and %85.56, respectively. Wu et al. (Wu, Wu, and Lee 2014) proposed a model based on the AprioriAll algorithm and k-means. They converted stock data into a set of charts that used a sliding window. These charts were then subdivided into clusters by the K-mean algorithm and achieved promising results.

## *1.4 Sentiment Analysis*

Sentiment Analysis (SA) is an NLP task that attempts to classify a document into a set of polar classes (negative, positive, or neutral) (Zhou and Zafarani). Generally, SA is intended to determine the attitude of a speaker or a writer in relation to a subject, which is generally expressed as a non-structural text (Rapoza). SA has become a big part of the stock market, as the idea of SA based on different data sources can provide insights into how stock markets respond to different types of news in the short and medium time. (Shah, Isah, and Zulkernine 2019) proposed a method that measures sentiment from sources or the sentiment behind the news to determine its impact on stock markets—also, Lee et al.(Lee et al. 2014) presented an approach to determine the importance of SA in stock market forecasting. The authors developed a system to predict whether stock prices will stay high or low by SA in the Form 8-K reports of their respective stocks. Cakra and Trisedya (Cakra and Trisedya 2015) attempted to forecast the price, price fluctuation, and margin percentage of Indonesian stocks using a simple SA model with classification techniques and a predictive model of linear regression. In this approach, the authors classified the tweets into three categories: positive, negative, and neutral, and then ignored the neutral tweets as promotional tweets or spam tweets. Their experiments showed that the Random Forest algorithm performed the best against other classifiers with 60.39% accuracy.

## 1.5 Hybrid Approaches

Hybrid approaches try to combine different learning approaches to create a better learning model. For example, Hyun-jung Kim and Kyung-shik Shin (Kim and Shin 2007) presented a hybrid approach based on the artificial neural networks (ANNs) for time series property extraction. They separately combined the adaptive time-delay neural networks (ATNNs) and the time delay neural networks (TDNNs), with the genetic algorithms (GAs). Their proposed model tries to detect temporal patterns for stock market prediction tasks. The results show that the mean square error (MSE) of the derived GA–ATNN and GA–TDNN from the genetic search process is smaller than that of the standard ATNN, TDNN, and RNN. Ding et al. (Ding et al. 2015)used a novel neural tensor network to model the effects of long-term and short-term events on stock price movement using event embedding and Convolutional Neural Networks(CNNs). Compared to the other approaches, the embedded representation of events was able to achieve higher accuracy.

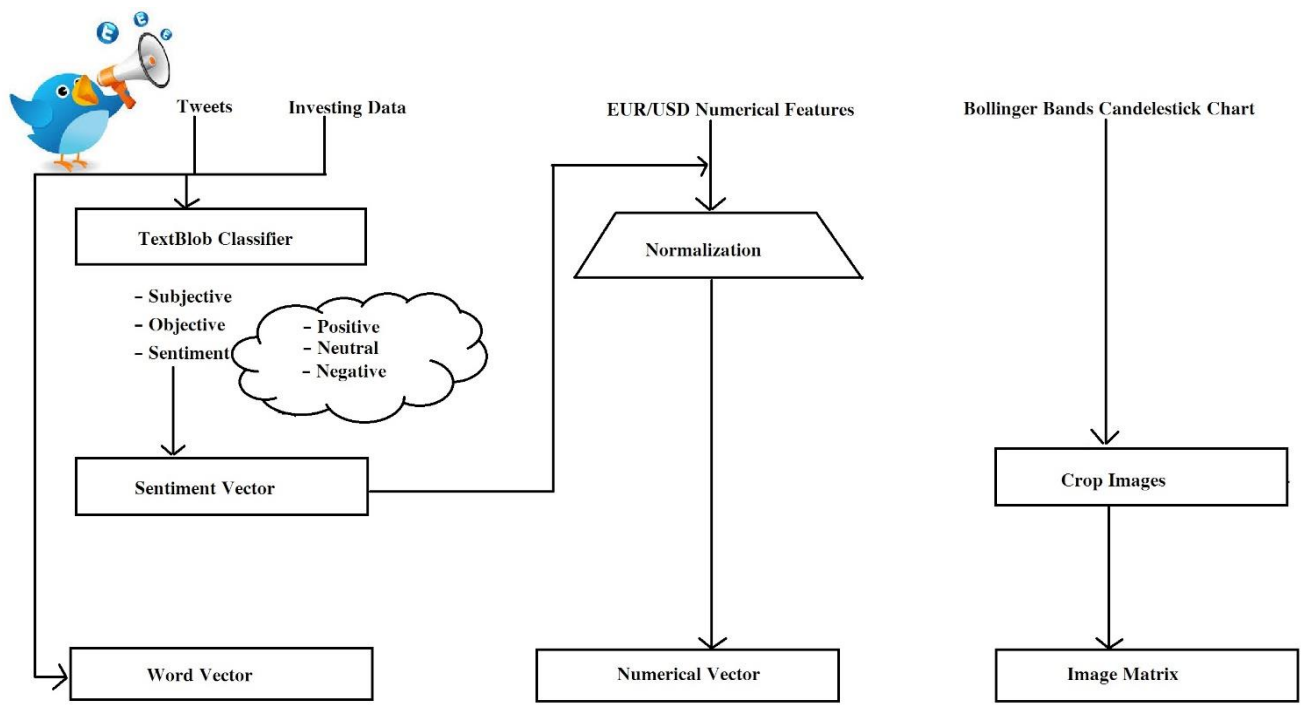

Figure 1: Flowchart of the proposed methodology for web data mining. (Word Vector) results for twitter and investing words for EUR/USD stock, (Numerical Vector) includes Objective, Subjective, Sentiment, Price, Open, High, Low, and Exchange stock for EUR/USD, and (Image Matrix) cropping candlestick charts.

## 2 Web Mining and Preliminary Data Processing

In this section we describe the process of mining and collecting data to create the vectors and matrix specified in Figure 1, mainly the Word Vector, the Numerical Vector, and the Image Matrix.

### 2.1 Word Vector

Before considering how to create this vector, let's look at how to collect textual data. Searching for information on Twitter is based on hashtags. Various implementation tools have been made available to the public. GetOldTweets3[1] has been used in current work to collect data from Twitter., which is a completely free tool for collecting Twitter data that also supports hybrid search and word search features, allowing long term tweets. The hashtag #EUR/USD was used to search using this tool. 56,954 of tweets were extracted, which was reduced to 25,793 after pre-processing. The pre-processing steps are as follows:

- Language detection: In this phase, some tweets that have been collected in non-English languages will be deleted. For this purpose, the word count rate, a common technique for identifying a language, was used.
- Stop-words Removal: Using the NLTK[2] library, all the stop-words in the tweets were deleted.
- URLs and Punctuations Removal: All URL links and punctuations are removed because they do not affect the final accuracy. By eliminating these, many tweets became None.
- None Record Removal: The None tweets are deleted in this step.

To offset the contextual data related to the EUR/USD stock indices, the information available on the Investing web site (announced news) was collected. Python's Request[3] library was used for this task. Similar pre-processing steps (to Tweeter data) have been performed on this textual data, except the step 1 above. A Word Vector is created for each comment (sequence of words) extracted from the Tweeter or Investing web site. The Word Vector can be represented by $V = [V_1, V_2, ..., V_n]$, where $V_i$ refers to the word $i$ in this sequence.

*2.2 Numerical Vector*

As shown in Figure 1, this vector is created by a combination of sentiment and the EUR/USD numerical features finance vectors. The sentiment vector is the result of applying the TextBlob[4] tool to comments posted on Twitter or news on the Investing web site, which results in the three numerical properties of Objective, Subjective, and Sentiment. The subjective and objective properties are numbers between 0 and 1 with a sentiment property rated as 0-1, assigned to each. The Numerical Vector shows the daily changes in the EUR/USD stock market. Since the purpose of this project is to identify the features that affect the increase/decrease of the EUR/USD stock market, it is necessary to collect attribute values on a daily basis. Online trading and momentum price statistics of the EUR/USD stock market are provided by several free web sites and API, including Google, Forex factory, and Yahoo finance. However, the statistics are publicly released with a 20-minute delay. In addition to daily snapshots, some sources on the Internet provide a history of this information, most notably Yahoo and Google. Although the data from these two web sites are valid, unfortunately, their data only contains daily data and does not cover day-to-day price changes. One of the most popular sources that publish the daily history of shared data is the forex factory1 web site that provides the values for attributes of Price, Open, High, Low, and Exchange, which are useful in indicating a decrease or increase in data. In current work, the Exchange value is mapped to two numbers 0 (if negative) and 1(if positive). This makes it easier to create a binary classifier.

---

[1] https://pypi.org/project/GetOldTweets3/
[2] https://www.nltk.org/
[3] https://requests.readthedocs.io/en/master/
[4] https://textblob.readthedocs.io/en/dev/

### 2.3 *Image matrix*

The image matrix contains visual information of daily Candlestick charts of stock changes. Candlestick charts are one about the most important charts on the stock market to show the extent of changes over time. Steve Nisson (Nison 1994) was the first to provide a graphical profile of these charts. Schlossberg (Lien 2008) proved that Candlesticks alone can almost be "useless" and that their full potential is reached when used in conjunction with other technical indicators. Candlestick charts are made with four variables: open, closing, high, and low over the period. The visual appeal and information contained in each candlestick are what make this diagram so useful for data mining and data analysis. Windsor Brokers2 was used to extract the candlestick charts, which provide daily information on the EUR/USD stock market and their history. These charts were extracted daily from Jan 2nd 2007 to 7th Jun 2019. This tool enables the extraction of charts in different viewpoints such as Bollinger Bands, Heikin-Ashi, etc. Bollinger Bands' perspective was used in this study. Figure 2 is an example of a chart extracted by this tool. These Candlestick charts are considered as image matrices in the dimensions specified for the network.

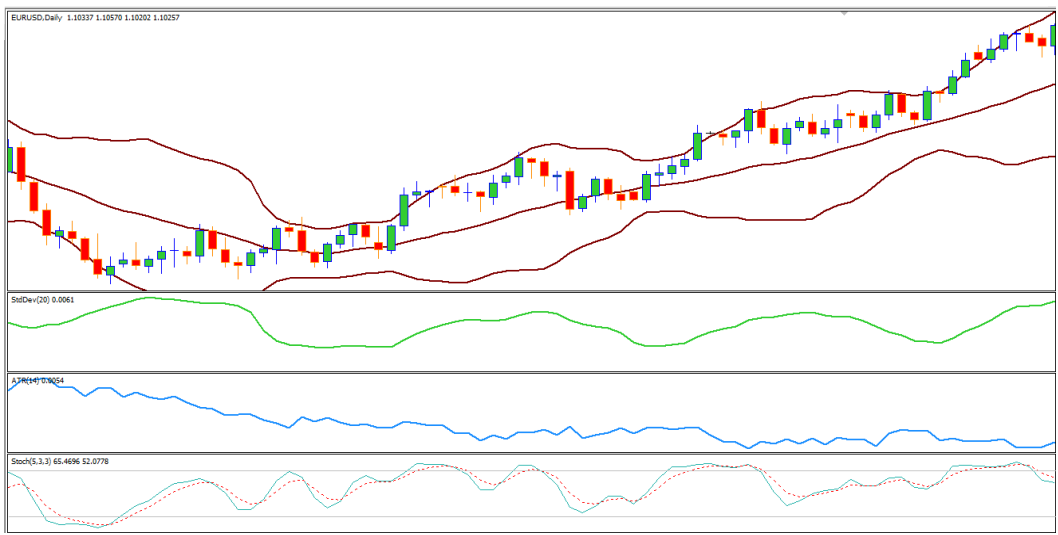

Figure 2: EUR/USD, Jan 2nd 2007, Bollinger Bands.

---

1 https://www.forexfactory.com
2 https://en.windsorbrokers.com/

## 3 Experimental setup

A brief explanation of some tools used in the proposed models and evaluation measures is introduced in the following subsections.

### 3.1 Glove word embedding

One of the most important issues that should be taken into account when working with the neural network on the text, is that cannot directly feed the raw text to the neural network, as the neural network receives D-dimensional feature vectors. One-hot encoding is not suitable due to the length of the dictionary size. so, there is a need to embed each feature into a D-dimensional space and represent it as a dense vector in the space. The solution is called word embeddings, which creates a single D-dimensional dense vector for each feature. The vectors are very flexible and help avoiding the curse of dimensionality. Generally, for creating dense vectors, the word embedding methods are trained on a large volume dataset. Word2vec (Mikolov et al. 2013) and Glove (Pennington, Socher, and Manning 2014) are two popular methods to create these dense vectors. Pre-trained models are also very common. The most important of which are the pre-trained word2vec-GoogleNews-vectors1 model and the pre-trained Glove2 model. In the present study, the pre-trained Glove was used to create dense vectors. That includes 1.2M vocab.

### 3.2 Keras deep learning library

The Keras[3] is a high-level library for creating neural network tools written in the Python and capable of running on TensorFlow4, CNTK5, and Theano6. Keras includes several implementations of neural network structure blocks, such as layer, objection, activation function, and optimizer, as well as numerous tools for images and text data. The Keras has been used in current work for developing the TI-Capsule model described in section 5.

### 3.3 evaluation measures

The quality of a classifier can be represented by a confusion matrix. Each of its records represents the correct and incorrect example for each class. For example, Table 1 is an example of this matrix for a binary classifier:

Table 1: a confusion matrix for binary classifier

| class | classified | |
|---|---|---|
| | As pos | As neg |

---

[1] https://github.com/mmihaltz/word2vec-GoogleNews-vectors
[2] https://nlp.stanford.edu/projects/glove/
[3] https://keras.io
[4] https://www.tensorflow.org/
[5] https://cntk.ai/pythondocs/
[6] https://pypi.org/project/Theano/

|      |    |    |
|------|----|----|
| pos  | TP | FN |
| neg  | FP | TN |

On the other hand, the efficiency of a classifier can be determined by applying it to test data, but accuracy is not the only measure to evaluate this performance their various measure that can be used to evaluate the efficiency of a classifier. In this paper, we use a measure such as Accuracy, Precision, Recall, and F1_score to evaluate the proposed model.

$$Precision = \frac{TP}{TP + FP} \qquad (1)$$

$$Recall = \frac{TP}{TP + FN} \qquad (2)$$

$$F_1 = 2 * \frac{Precision * Recall}{Precision + Recall} \qquad (3)$$

$$Accuracy = \frac{TP + TN}{TP + TN + FP + FN} \qquad (4)$$

TP is the number of positive samples classified as positive, FP number of negative samples classified as positive, TN is the number of negative samples classified as negative, and FN is the number of positive samples classified as negative.

## 4 The Model Architecture

This section describes the structure of the proposed TI-Capsule in detail. The proposed approach actually consists of two parallel capsule grids that extract the latent features from the texts and images. The two main branches in Figure 3 demonstrate these two paths for text and image latent feature extraction. These features are merged into a new space that represents both image and text inputs. First, a brief look at the capsule networks is provided.

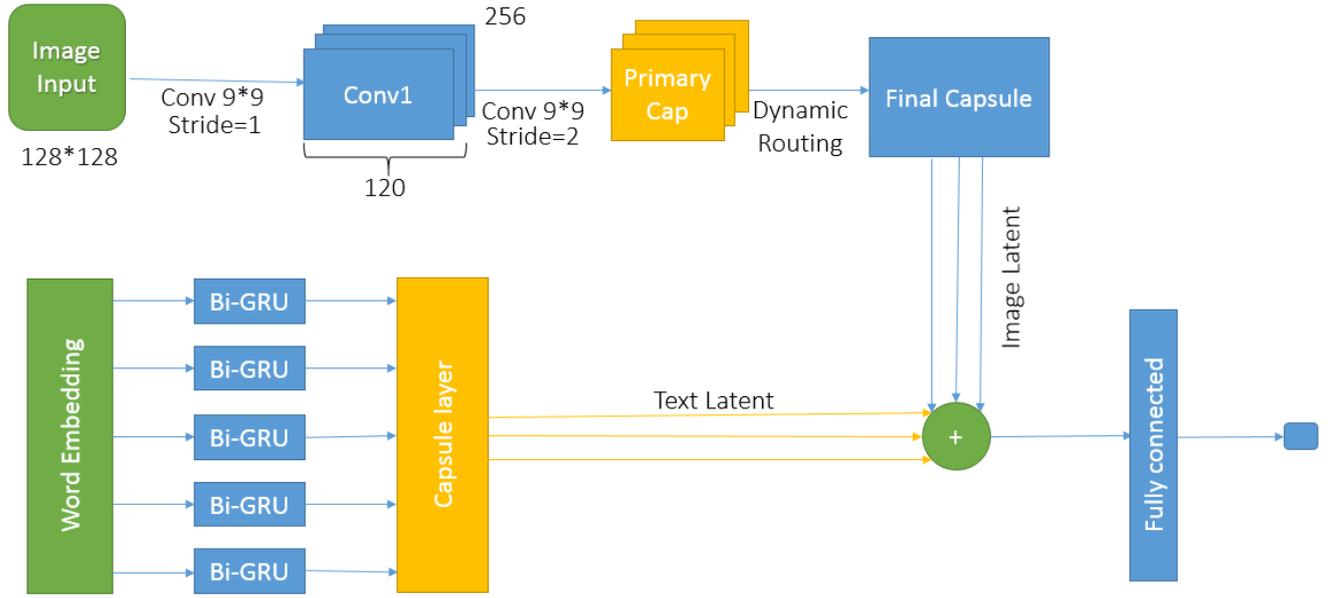

Figure 3: The Architect of proposed model

## 4.1 The Capsule network

In recent years Capsule networks have gained great efficiency for various tasks such as image recognition (Abdel-Hamid et al. 2014), and NLP (Lopez and Kalita 2017). Capsule networks were first introduced by Sabour et al (Sabour, Frosst, and Hinton 2017). The network tries to store information in sorted vectors instead of scaler values. The vectors contain various information such as spatial orientation, magnitude/prevalence, and other attributes of the extracted features, which are represented by different layers of the Capsule network. The set of neurons are considered as a vector. In this network, instead of sending scalar values to the next layer, the capsule's output in one layer are transferred to the next layer. Such transformation can be carried out by an algorithm like Dynamic Routing (Sabour, Frosst, and Hinton 2017), which is an agreement between the two high-level (the next layer) and the low-level capsules (the current layer) usage.

## 4.2 Text branch

The purpose of the text branch in Figure 3 is to extract a set of latent attributes that are denoted by $X^{tl}$. To extract these features, a Bi-directional GRU Capsule network (Khayi and Rus) is created, which consists of three basic layers: Word Embedding layer, Bi-directional GRU layer, and Capsule layer.

1. **Word Embedding layer:** In this layer, each document (opinion/news) is converted into a set of dense vectors. Let denote the $j-th$ word in the document (opinion/news) as $X^{ij} \in R^k$, which is a k-dimensional word embedding vector. Pre-trained 100-dimensional Glove vector values have been used to create these dense vectors[1].

2. Due to the limitation of the input length of the Bi-GRU networks and the different length of the input documents, the maximum document lengths in the whole dataset is considered as the threshold. Suppose the maximum length of documents in the dataset is $n$, so, the documents that have less than $n$ words can be padded with zero paddings as a sequence with length $n$. Hence, the overall documents can be written as:

$$X_{i,l:n}^{tl} = X^{i,1} \oplus X^{i,2} \oplus \ldots \oplus X^{i,n} \tag{5}$$

where $\oplus$ is the concatenation operator. The output of this layer is a matrix representation for each document as follows:

$$out_{embd} = N * V \tag{6}$$

Where $N$ represents the maximum length of the documents in the dataset and $V$ represents the length of the dense vector.

3. **Bi-directional GRU layer**: Initial features for capsule networks should be extracted by convolution or recurrent layers. Because of the nature of the data in current work, which is in the form of time series, the Bi-directional version of the GRU network (Cho et al. 2014) is used. The output of the Word embedding layer is considered as the input for this layer. There is a forward GRU ($\overrightarrow{h_t}$) and backward GRU ($\overleftarrow{h_t}$) in this layer, which receives the input and the reverse of the input, respectively. The final output comes from the integration of forward and backward ($\overrightarrow{h_t}, \overleftarrow{h_t}$) GRUs. Based on the document representations which is explained above ($X_{i,l:n}^{tl} = X^{i,1} \oplus X^{i,2} \oplus \ldots \oplus X^{i,n}$), the $ht = [h1, h2, \ldots, ht]$ or the hidden vectors sequence in GRU is calculated by following equations:

$$z_t = \sigma(w_z x_t + U_z h_{t-1} + b_z) \tag{7}$$
$$r_t = \sigma(w_r x_t + U_r h_{t-1} + b_r) \tag{8}$$
$$h'_t = \sigma(w_h x_t + U_n(r_t \odot h_{t-1}) + b_h) \tag{9}$$
$$h_t = (1-z)h_{t-1} + z_t h'_t \tag{10}$$

Where $z_t$ is the update gate, $r_t$ is the rest gate, $h'_t$ is the candidate gate, and $h_t$ is the output vector. $[W_Z, W_R, W_N, U_Z, U_R, U_N]$ are learnable matrixes and $[b_n, b_z, b_r]$ are learnable biases. $\sigma$ is sigmoid activation function, and $\odot$ is an elementwise multiplication.

4. Capsule layer: Bi-GRU's encoded features are then given to a Capsule layer. This network includes a set of capsules. Each capsule corresponds to a high-level feature. If Bi-GRU output is shown by $h_i$, and $w$ is a weighted matrix, then $\hat{v}_{i|j}$, which represents the predictor vector, is obtained from the following equation:

$$\hat{v}_{i|j} = w_{ij} h_i \tag{11}$$

---

[1] http://nlp.stanford.edu/data/glove.twitter.27B.zip

Where $w_{ij}$ is learnable matrix. The given value to the capsule $S_j$ contains the weighted sum of all the prediction vectors $\hat{v}_{i|j}$ which is computed according to the equation:

$$S_j = \sum_i c_{ij}\,\hat{v}_{i|j} \tag{12}$$

Where $c_{ij}$ is the coupling coefficient, which is repeatedly adjusted by Dynamic Routing algorithm (Sabour, Frosst, and Hinton 2017). Finally, the latent properties of the text are obtained from the flattened value of $v_j$. The "squash" is used as a non-linear function for mapping the values of $S_j$ vectors to obtain $v_j$. This function is applied to $S_j$ according to the following equation.

$$v_j = \frac{||s_j||^2}{1 + ||s_j||^2} \cdot \frac{s_j}{||s_j||} \tag{13}$$

### *4.3 Image branch*

Similar to the text branch, a capsule network is used to extract the latent image features which are called $X^{il}$. A convolution network is used to extract basic image features, as generally convolutional networks are shown promising results for image feature extractions. Due to the size limitation of the convolution layer, a 128 * 128 image size was considered for the input images. Size selection has a great impact on processing speed as well.

The first layer of the Image branch (Figure 3) is a convolution layer with a filter size of 256*9*9 and stride=1 (referred to as Conv1). Depending on the used filters, it will produce 256 different feature maps with a size of 120 *120. In this layer, ReLu was also used as an activation function.

The second layer is a primary capsule that attempts to find small parts of the Conv1 layer properties as input and extract specific patterns according to them. This layer maintains the local order between the input image components (referred to as PrimeryCaps). This layer contains 32 channels of 8-dimensional capsules, which means that each capsule is a vector made up of 8 convolutional neurons with a 9 *9 filter as well as a stride=2. The primary capsule layer possesses $[32 * 56 * 56]$ capsule outputs.

The final layer is the Final Capsule (referred to as FinalCaps). The PrimeryCaps layer attempts to extract the low-level features extracted by the Conv1 layer. The FinalCaps layer is a higher level that used to learn and display more complex information, such as a class of data. The dimensions of this layer can be equal to the data class or different. In this layer, 10 capsules are considered, and the dimension of these capsules is set to 16. The Flatten value of this layer is considered to be Image branch latent features.

At the end, the two branches were assembled into a fully connected layer for classification.

## 5    Experimental Results

In our experiments 80% of the data has been used for training, and 20% of the data has been used for testing. Since, there are 4 different vectors (Word Vector, Numerical Vector, Sentiment Vector, and Finance Vector) and a matrix (Image Matrix) of data in the proposed model, when it comes to choose the best parameters for each model, the number of results increases exponentially.

For sensitivity analysis, the effectiveness of two important parameters has been studied on the proposed TI-Capsule model that are the batch size and the hidden layer dimensions.

- **Batch size:** The batch size represents the number of data samples that are going to be propagated through the network. The larger this parameter the network consumes more memory, and the smaller the value, the training time of the network is longer. For this purpose, different values of 8, 16, 32, 64, 128, and 256 have been evaluated for this parameter. Figure 4 shows the effect of batch size on evolution metrics for the proposed TI-Capsule model. Based on figure 4, the best value of 8 is selected for the final model in our experiment.

- **Hidden layer dimension:** Deciding on the number of neurons in the hidden layers is an important part of the overall architecture of neural networks. These layers are actually responsible for learning the features extracted by other layers of the network. Using very few neurons in the hidden layers leads to underfitting problem and oppositely using too many neurons in the hidden layers can also cause several problems such as overfitting. Therefore, a balanced number of neurons must be used in the layers. So, different number of neurons including 8, 16, 32, 64, and 128 and 256 has been examined in our experiments. Figure 5 shows the effect of Hidden layer dimension on evolution metrics for the proposed TI-Capsule model. Based on figure 5, the best value of 8 is selected for the final model in our experiment.

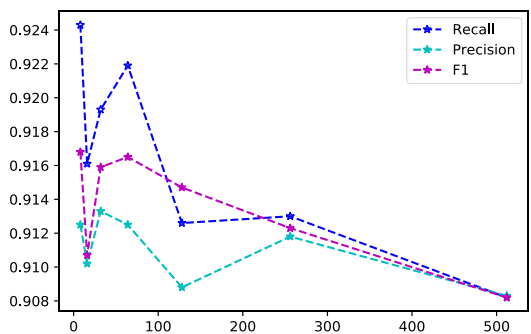 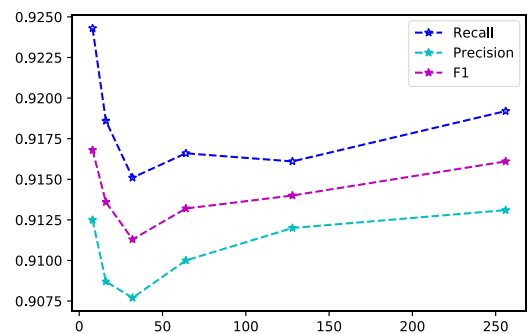

Figure 4: Batch size impact on evaluation metrics.    Figure 5: Hidden *layer* dimension impact on evaluation metrics.

A comparison has been performed among our proposed TI-Capsule model with the six baseline models that are detailed below (the results explained in section 7). These models are derived from the architectures of LSTM, GRU, Bidirectional-GRUCapsule and Multi-channel CNN.

LSTM-features-4: In this model, four modified scale properties, Price, Open, High, and Low, are given as inputs. The purpose of this model is to find the effect of numerical features on the final prediction.

1. **LSTM-features-7:** In this model, seven modified scale properties, Price, Open, High, Low, subjective, objective and, text sentiment, are given as inputs. The purpose of this model is to find the effect of numerical features with sentiment features on the final prediction. In both LSTM-features-4 and LSTM-features-7 models, a single layer of LSTM was used.
2. **LSTM-text-71:** In this model the maximum length of tweets and news was chosen as the threshold for padding, which is 71. So, word strings of length 71 are received as input. The 100-dimensional Glove word embedding were also considered as initial weights in this model.
3. **GRU-text-71:** This model is similar to LSTM71 except that GRU blocks are used instead of LSTM blocks.
4. **Multi-channel CNN-text-71:** This model is similar to the multi-channel CNN approach introduced in (Kim 2014). It has four layers of convolution, max-pooling, dropout, and fully-connected.
5. **Bidirectional-GRUCapsule-text-71:** Our Bi-GRUCapsule model is similar to the Bi-GRUCapsule approach introduced in (Kim 2014). This approach consists of 4 different layers of word-embedding, Bi-GRU, Capsule, and classification layer. All four LSTM-text-71, GRU-text-71, Multi-channel CNN-text-71, and Bidirectional-GRUCapsule-text-71 models are trained to investigate the impact of contextual data on the final prediction.

As mentioned in some of the above models, the number after the model names is the maximum length of input data.

Table 2 shows the results of applying the 6 different baseline models together with the proposed TI-Capsule model on the target dataset. As it is clear from the table, the two LSTM-features-4 and LSTM-features-7 models trained only on numerical features. They have the worst accuracy because the existence of numerical features alone is not sufficient for prediction. Among the text-based approaches, Bidirectional-GRUCapsule-text-71 approach has the worst results, while Multi-channel CNN-text-71 is the most accurate. Using text, feature, and image information, TI-Capsule outperforms all the baseline methods significantly.

Table 2: Result obtained by 6 baseline models and TI-Capsule approach.

| Method | Accuracy | Precision | Recall | F1-measure |
|---|---|---|---|---|

| LSTM-features-4 | 0.5753 | 0.5909 | 0.3714 | 0.4561 |
| LSTM-features-7 | 0.5873 | 0.6062 | 0.8238 | 0.6980 |
| LSTM-text-71 | 0.8073 | 0.7837 | 0.9272 | 0.8485 |
| GRU-text-71 | 0.8048 | 0.7903 | 0.9074 | 0.8436 |
| Multi-channel CNN-text-71 | 0.8922 | 0.8814 | **0.9418** | 0.9101 |
| Bidirectional-GRUCapsule-text-71 | 0.7819 | 0.7728 | 0.8881 | 0.8253 |
| **TI-capsule** | **0.9126** | **0.9120** | 0.9161 | **0.9140** |

The loss rate of applying the 6 baseline models together with the proposed TI-Capsule model on the data set are shown in Figures 6. The loss rate has been reported on the data after 50 times running of each model.

Figure 6-1 shows the error related to the LSTM-features-4 model. This model has a logarithmic mode, which shows that in each epoch, training and testing errors were almost reduced. Also, the short distance between training and testing indicates the absence of overfitting in the model. Figure 6-2 shows the error related to the LSTM-features-7 model and this error has a lot of fluctuations that the basis of this fluctuation is due to not adjusting the parameters of the model. Also, the error on the training data is reduced much better than the test data, which tends to overfitting at higher epochs.

Figure 6-3 and Figure 6-4 show the LSTM-features-4 and LSTM-features-7 models on the text data. These models have almost the same error pattern because they use the same time series nature and the same inputs. The distance between training and test error in this diagram is very small, which indicates the absence of overfitting. The two LSTM-features-4 and LSTM-features-7 approaches have the most errors in the training and testing phase, which indicates that numerical properties alone cannot be useful. In the LSTM-features-7 model, the error rate is roughly declining, while the LSTM-features-4 model has a more regular downward trend. In text-based models, Multi-channel CNN-text-71 obtained the least loss (Figure 6-5 shows the error related to this model). However, this model tends to overfitting as the number of epochs increases. The three LSTM-text-71, GRU-text-71, and Bidirectional-GRUCapsule-text-71 models have the same downward trend due to having recursive layers. The TI-Capsule model achieves a lower loss than the all models (Figure 6-6 shows the error related to this model). On the other hand, the loss variations on the train and test data are very similar and no overfitting occurred.

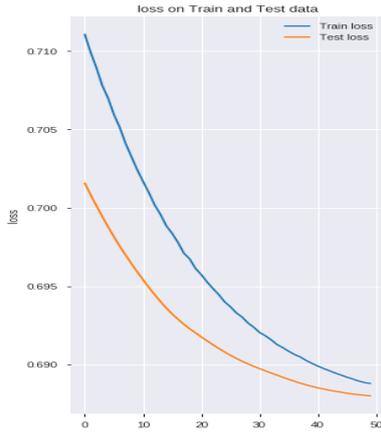
Figure 6-1: LSTM-features-4 loss

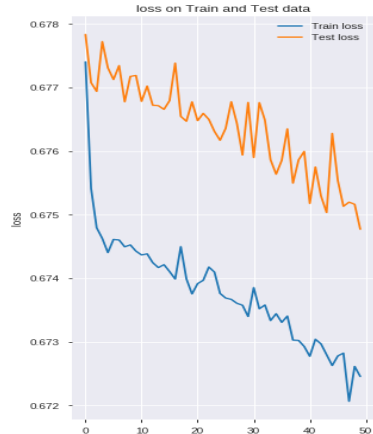
Figure 7-2: LSTM-features-7 loss

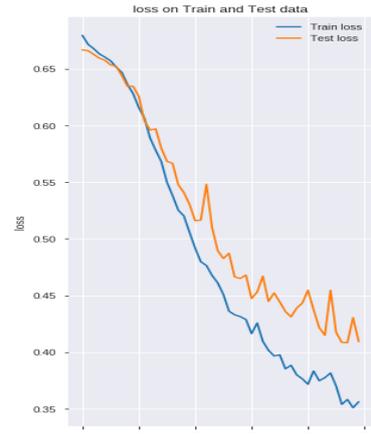
Figure 8-3: LSTM-text-71 loss

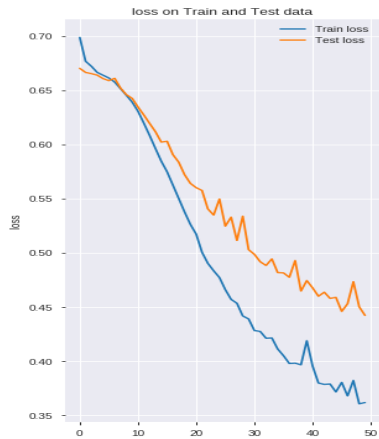
Figure 9-4: GRU-text-71 loss

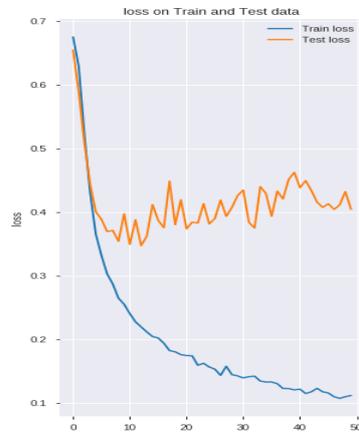
Figure 10-5: Multi-channel CNN-text-71 loss

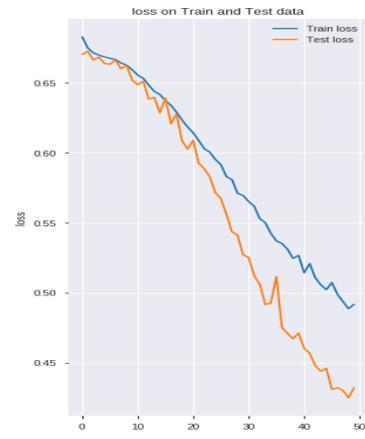
Figure 11-6: Bidirectional-GRUCapsule-text-71 loss

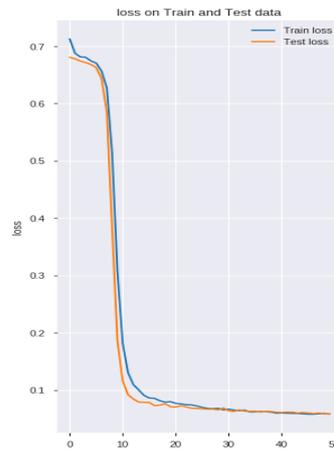
*Figure 12-7: TI-Capsule loss*

Figure 6: Loss obtained by the proposed approaches on the train and test data after 50 epochs.

## *6* **Conclusion**

Various researches have been performed to predict stock sentiment. Most studies use stock financial records and other market indices to predict stock sentiment using neural networks or other machine learning techniques. Some other methods use sentiment analysis to predict whether they are positive or negative. A number of studies also identify influencers on the social network and predict their views.

The purpose of this study, is to predict stock sentiment using multi-modal numerical and contextual data together with visual information on stock indices. Different models have been applied to different features of the target dataset. Numerical models (LSTM-features-4 and LSTM-features-7) have gained less accuracy in comparison to other models, which shows the existence of numerical data alone is not the solution. The Multi-channel CNN-text-71 model is able to achieve high accuracy on the textual data, but this model tended to overfitting in high epochs. The TI-Capsule approach gains the highest accuracy measure than the all other models. The loss diagram of this model shows that overfitting did not occur at higher epochs as well. This reveals that the integration of various data types including numerical, text and image data can lead to better prediction in stock market. One of the most important reasons that the proposed approach has been able to achieve a good result is the combination of visual and textual features. Capsule networks in the proposed approach maintain spatial relationships in images and semantic relationships in textual data well. Capsules in this model generate hidden beams of input data, which leads to more accurate outputs than traditional machine learning approaches and non-capsular deep learning approaches.

**References**

Abdel-Hamid, Ossama, Abdel-rahman Mohamed, Hui Jiang, Li Deng, Gerald Penn, and Dong Yu. 2014. 'Convolutional neural networks for speech recognition', *IEEE/ACM Transactions on audio, speech, and language processing*, 22: 1533-45.

Arévalo, Rubén, Jorge García, Francisco Guijarro, and Alfred Peris. 2017. 'A dynamic trading rule based on filtered flag pattern recognition for stock market price forecasting', *Expert Systems with Applications*, 81: 177-92.

Ariyo, Adebiyi A, Adewumi O Adewumi, and Charles K Ayo. 2014. "Stock price prediction using the ARIMA model." In *2014 UKSim-AMSS 16th International Conference on Computer Modelling and Simulation*, 106-12. IEEE.

Ballings, Michel, Dirk Van den Poel, Nathalie Hespeels, and Ruben Gryp. 2015. 'Evaluating multiple classifiers for stock price direction prediction', *Expert Systems with Applications*, 42: 7046-56.

Bartov, Eli, Lucile Faurel, and Partha S Mohanram. 2018. 'Can Twitter help predict firm-level earnings and stock returns?', *The Accounting Review*, 93: 25-57.

Bernal, Armando, Sam Fok, and Rohit Pidaparthi. 2012. 'Financial Market Time Series Prediction with Recurrent Neural Networks', *State College: Citeseer.[Google Scholar]*.

Billah, Baki, Maxwell L King, Ralph D Snyder, and Anne B Koehler. 2006. 'Exponential smoothing model selection for forecasting', *International Journal of Forecasting*, 22: 239-47.

Bollen, Johan, Huina Mao, and Xiaojun Zeng. 2011. 'Twitter mood predicts the stock market', *Journal of computational science*, 2: 1-8.

Bollerslev, Tim. 1986. 'Generalized autoregressive conditional heteroskedasticity', *Journal of econometrics*, 31: 307-27.

Cakra, Yahya Eru, and Bayu Distiawan Trisedya. 2015. "Stock price prediction using linear regression based on sentiment analysis." In *2015 international conference on advanced computer science and information systems (ICACSIS)*, 147-54. IEEE.

Cervelló-Royo, Roberto, Francisco Guijarro, and Karolina Michniuk. 2015. 'Stock market trading rule based on pattern recognition and technical analysis: Forecasting the DJIA index with intraday data', *Expert Systems with Applications*, 42: 5963-75.

Chen, Jiann-Fuh, Wei-Ming Wang, and Chao-Ming Huang. 1995. 'Analysis of an adaptive time-series autoregressive moving-average (ARMA) model for short-term load forecasting', *Electric Power Systems Research*, 34: 187-96.

Chen, Tai-liang, and Feng-yu Chen. 2016. 'An intelligent pattern recognition model for supporting investment decisions in stock market', *Information Sciences*, 346: 261-74.

Chen, Yubo, and Jinhong Xie. 2008. 'Online consumer review: Word-of-mouth as a new element of marketing communication mix', *Management science*, 54: 477-91.

Cho, Kyunghyun, Bart Van Merriënboer, Caglar Gulcehre, Dzmitry Bahdanau, Fethi Bougares, Holger Schwenk, and Yoshua Bengio. 2014. 'Learning phrase representations using RNN encoder-decoder for statistical machine translation', *arXiv preprint arXiv:1406.1078*.

De Faria, EL, Marcelo P Albuquerque, JL Gonzalez, JTP Cavalcante, and Marcio P Albuquerque. 2009. 'Predicting the Brazilian stock market through neural networks and adaptive exponential smoothing methods', *Expert Systems with Applications*, 36: 12506-09.

de O. Carosia, Arthur E, Guilherme P Coelho, and Ana EA da Silva. 2019. "The influence of tweets and news on the brazilian stock market through sentiment analysis." In *Proceedings of the 25th Brazillian Symposium on Multimedia and the Web*, 385-92.

Ding, Xiao, Yue Zhang, Ting Liu, and Junwen Duan. 2015. "Deep learning for event-driven stock prediction." In *Twenty-fourth international joint conference on artificial intelligence*.

Fu, Tak-chung, Fu-lai Chung, Robert Luk, and Chak-man Ng. 2005. "Preventing meaningless stock time series pattern discovery by changing perceptually important point detection." In *International Conference on Fuzzy Systems and Knowledge Discovery*, 1171-74. Springer.

Hamed, AL-Rubaiee, Renxi Qiu, and Dayou Li. 2015. "Analysis of the relationship between Saudi twitter posts and the Saudi stock market." In *2015 IEEE Seventh International Conference on Intelligent Computing and Information Systems (ICICIS)*, 660-65. IEEE.

Izzah, Abidatul, Yuita Arum Sari, Ratna Widyastuti, and Toga Aldila Cinderatama. 2017. "Mobile app for stock prediction using Improved Multiple Linear Regression." In *2017 International Conference on Sustainable Information Engineering and Technology (SIET)*, 150-54. IEEE.

Kamble, Rupesh A. 2017. "Short and long term stock trend prediction using decision tree." In *2017 International Conference on Intelligent Computing and Control Systems (ICICCS)*, 1371-75. IEEE.

Kaur, Ramandeep, and Sandeep Kautish. 2019. 'Multimodal Sentiment Analysis: A Survey and Comparison', *International Journal of Service Science, Management, Engineering, and Technology (IJSSMET)*, 10: 38-58.

Kharde, Vishal, and Prof Sonawane. 2016. 'Sentiment analysis of twitter data: a survey of techniques', *arXiv preprint arXiv:1601.06971*.

Khayi, Nisrine Ait, and Vasile Rus. 'BI-GRU Capsule Networks for Student Answers Assessment'.

Kim, Hyun-jung, and Kyung-shik Shin. 2007. 'A hybrid approach based on neural networks and genetic algorithms for detecting temporal patterns in stock markets', *Applied Soft Computing*, 7: 569-76.

Kim, Karam, Doojin Ryu, and Heejin Yang. 2019. 'Investor sentiment, stock returns, and analyst recommendation changes: The KOSPI stock market', *Investment Analysts Journal*, 48: 89-101.

Kim, Sang Hyuk, Hee Soo Lee, Han Jun Ko, Seung Hwan Jeong, Hyun Woo Byun, and Kyong Joo Oh. 2018. 'Pattern matching trading system based on the dynamic time warping algorithm', *Sustainability*, 10: 4641.

Kim, Yoon. 2014. 'Convolutional neural networks for sentence classification', *arXiv preprint arXiv:1408.5882*.
Lee, Heeyoung, Mihai Surdeanu, Bill MacCartney, and Dan Jurafsky. 2014. "On the Importance of Text Analysis for Stock Price Prediction." In *LREC*, 1170-75.
Li, Qian, Bing Zhou, and Qingzhong Liu. 2016. "Can twitter posts predict stock behavior?: A study of stock market with twitter social emotion." In *2016 IEEE International Conference on Cloud Computing and Big Data Analysis (ICCCBDA)*, 359-64. IEEE.
Lien, Kathy. 2008. *Day trading and swing trading the currency market: technical and fundamental strategies to profit from market moves* (John Wiley & Sons).
Lopez, Marc Moreno, and Jugal Kalita. 2017. 'Deep Learning applied to NLP', *arXiv preprint arXiv:1703.03091*.
Maerz, Seraphine F, and Cornelius Puschmann. 2020. 'Text as Data for Conflict Research: A Literature Survey.' in, *Computational Conflict Research* (Springer).
Mikolov, Tomas, Kai Chen, Greg Corrado, and Jeffrey Dean. 2013. 'Efficient estimation of word representations in vector space', *arXiv preprint arXiv:1301.3781*.
Nison, Steve. 1994. *Beyond candlesticks: new Japanese charting techniques revealed* (John Wiley & Sons).
Ouahilal, Meryem, Mohammed El Mohajir, Mohamed Chahhou, and Badr Eddine El Mohajir. 2016. "Optimizing stock market price prediction using a hybrid approach based on HP filter and support vector regression." In *2016 4th IEEE International Colloquium on Information Science and Technology (CiSt)*, 290-94. IEEE.
Pang, Bo, and Lillian Lee. 2008. 'Opinion mining and sentiment analysis', *Foundations and Trends® in Information Retrieval*, 2: 1-135.
Parracho, Paulo, Rui Neves, and Nuno Horta. 2010. "Trading in financial markets using pattern recognition optimized by genetic algorithms." In *Proceedings of the 12th annual conference companion on Genetic and evolutionary computation*, 2105-06.
Pennington, Jeffrey, Richard Socher, and Christopher D Manning. 2014. "Glove: Global vectors for word representation." In *Proceedings of the 2014 conference on empirical methods in natural language processing (EMNLP)*, 1532-43.
Phetchanchai, Chawalsak, Ali Selamat, Amjad Rehman, and Tanzila Saba. 2010. "Index financial time series based on zigzag-perceptually important points." In *Journal of Computer Science*. Citeseer.
Powell, Nicole, Simon Y Foo, and Mark Weatherspoon. 2008. "Supervised and unsupervised methods for stock trend forecasting." In *2008 40th Southeastern Symposium on System Theory (SSST)*, 203-05. IEEE.
Rao, Tushar, and Saket Srivastava. 2012. 'Analyzing stock market movements using twitter sentiment analysis'.
Rapoza, Kenneth %J Pridobljeno iz www. forbes. com/sites/kenrapoza//02/26/can-fake-news-impact-the-stock-market/. 2017. 'Can 'fake news' impact the stock market?'.
Ravi, Kumar, and Vadlamani Ravi. 2015. 'A survey on opinion mining and sentiment analysis: tasks, approaches and applications', *Knowledge-Based Systems*, 89: 14-46.
Sabour, Sara, Nicholas Frosst, and Geoffrey E Hinton. 2017. "Dynamic routing between capsules." In *Advances in neural information processing systems*, 3856-66.
Shad Akhtar, Md, Dushyant Singh Chauhan, Deepanway Ghosal, Soujanya Poria, Asif Ekbal, and Pushpak Bhattacharyya. 2019. 'Multi-task Learning for Multi-modal Emotion Recognition and Sentiment Analysis', *arXiv preprint arXiv:1905.05812*.
Shah, Dev, Wesley Campbell, and Farhana H Zulkernine. 2018. "A Comparative Study of LSTM and DNN for Stock Market Forecasting." In *2018 IEEE International Conference on Big Data (Big Data)*, 4148-55. IEEE.
Shah, Dev, Haruna Isah, and Farhana Zulkernine. 2019. 'Stock Market Analysis: A Review and Taxonomy of Prediction Techniques', *International Journal of Financial Studies*, 7: 26.

Sun, Zhongkai, Prathusha K Sarma, William Sethares, and Erik P Bucy. 2019. 'Multi-modal Sentiment Analysis using Deep Canonical Correlation Analysis', *arXiv preprint arXiv:1907.08696*.

Teräsvirta, Timo, Dick Van Dijk, and Marcelo C Medeiros. 2005. 'Linear models, smooth transition autoregressions, and neural networks for forecasting macroeconomic time series: A re-examination', *International Journal of Forecasting*, 21: 755-74.

Van Kleef, Ellen, Hans CM Van Trijp, and Pieternel Luning. 2005. 'Consumer research in the early stages of new product development: a critical review of methods and techniques', *Food quality and preference*, 16: 181-201.

Vu, Tien Thanh, Shu Chang, Quang Thuy Ha, and Nigel Collier. 2012. "An experiment in integrating sentiment features for tech stock prediction in twitter." In *Proceedings of the workshop on information extraction and entity analytics on social media data*, 23-38.

Wu, Kuo-Ping, Yung-Piao Wu, and Hahn-Ming Lee. 2014. 'Stock Trend Prediction by Using K-Means and AprioriAll Algorithm for Sequential Chart Pattern Mining', *J. Inf. Sci. Eng.*, 30: 669-86.

Xu, Feifei, and Vlado Keelj. 2014. "Collective sentiment mining of microblogs in 24-hour stock price movement prediction." In *2014 IEEE 16th Conference on Business Informatics*, 60-67. IEEE.

Yu, Lijing, Lingling Zhou, Li Tan, Hongbo Jiang, Ying Wang, Sheng Wei, and Shaofa Nie. 2014. 'Application of a new hybrid model with seasonal auto-regressive integrated moving average (ARIMA) and nonlinear auto-regressive neural network (NARNN) in forecasting incidence cases of HFMD in Shenzhen, China', *PloS one*, 9.

Zadrozny, Wlodek. 2019. "A Comparison of Neural Network Methods for Accurate Sentiment Analysis of Stock Market Tweets." In *ECML PKDD 2018 Workshops: MIDAS 2018 and PAP 2018, Dublin, Ireland, September 10-14, 2018, Proceedings*, 51. Springer.

Zhang, Dong, Shoushan Li, Qiaoming Zhu, and Guodong Zhou. 2019. "Effective Sentiment-relevant Word Selection for Multi-modal Sentiment Analysis in Spoken Language." In *Proceedings of the 27th ACM International Conference on Multimedia*, 148-56.

Zhang, Linhao. 2013. 'Sentiment analysis on Twitter with stock price and significant keyword correlation'.

Zhang, Xue, Hauke Fuehres, and Peter A Gloor. 2011. 'Predicting stock market indicators through twitter "I hope it is not as bad as I fear"', *Procedia-Social and Behavioral Sciences*, 26: 55-62.

Zhao, Bo, Yongji He, Chunfeng Yuan, and Yihua Huang. 2016. "Stock market prediction exploiting microblog sentiment analysis." In *2016 International Joint Conference on Neural Networks (IJCNN)*, 4482-88. IEEE.

Zhou, Xinyi, and Reza %J arXiv preprint arXiv:.00315 Zafarani. 2018. 'Fake news: A survey of research, detection methods, and opportunities'.

Zucco, Chiara, Barbara Calabrese, Giuseppe Agapito, Pietro H Guzzi, and Mario Cannataro. 2020. 'Sentiment analysis for mining texts and social networks data: Methods and tools', *Wiley Interdisciplinary Reviews: Data Mining and Knowledge Discovery*, 10: e1333.